\documentclass[letterpaper]{article} 
\usepackage{aaai2026}  
\usepackage{times}  
\usepackage{helvet}  
\usepackage{courier}  
\usepackage[hyphens]{url}  
\usepackage{graphicx} 
\usepackage{natbib}  
\usepackage{caption} 
\usepackage{algorithm}
\usepackage{algorithmic}

\usepackage{newfloat}
\usepackage{listings}
\DeclareCaptionStyle{ruled}{labelfont=normalfont,labelsep=colon,strut=off} 
\floatstyle{ruled}
\newfloat{listing}{tb}{lst}{}
\floatname{listing}{Listing}
\usepackage{booktabs}
\usepackage{amsmath}
\usepackage{amssymb}
\usepackage{graphicx}
\usepackage{subcaption}
\usepackage{xcolor}

\title{Learning to Generate and Extract: A Multi-Agent Collaboration Framework For Zero-shot Document-level Event Arguments Extraction}

\author {
     Guangjun Zhang\textsuperscript{\rm 1},
    Hu Zhang\textsuperscript{\rm 1,2}\thanks{corresponding author.},
    Yazhou Han\textsuperscript{\rm 1},
    Yue Fan\textsuperscript{\rm 1},
    Yuhang Shao\textsuperscript{\rm 1},
    Hongye Tan\textsuperscript{\rm 1,2},
    Ru Li\textsuperscript{\rm 1,2}
}
\affiliations {
     \textsuperscript{\rm 1}School of Computer and Information Technology, Shanxi University, Taiyuan, China\\
     \textsuperscript{\rm 2}Key Laboratory of Computational Intelligence and Chinese Information
 Processing of Ministry of Education, Shanxi University, Taiyuan, China \\
    zgj2866@gmail.com, \{zhanghu, hanyazhou, 202322408041, tanhongye, liru\}@sxu.edu.cn, yuefan24@163.com
}

\usepackage{bibentry}

\begin{document}

\maketitle

\begin{abstract}
Document-level event argument extraction (DEAE) is essential for knowledge acquisition, aiming to extract participants of events from documents. In the zero-shot setting, existing methods employ LLMs to generate synthetic data to address the challenge posed by the scarcity of annotated data. However, relying solely on \textit{Event-type-only prompts} makes it difficult for the generated content to accurately capture the contextual and structural relationships of unseen events. Moreover, ensuring the reliability and usability of synthetic data remains a significant challenge due to the absence of quality evaluation mechanisms.
To this end, we introduce a multi-agent collaboration framework for zero-shot document-level event argument extraction (ZS-DEAE), which simulates the human collaborative cognitive process of “Propose–Evaluate–Revise.” Specifically, the framework comprises a generation agent and an evaluation agent. The generation agent synthesizes data for unseen events by leveraging knowledge from seen events, while the evaluation agent extracts arguments from the synthetic data and assesses their semantic consistency with the context. 
The evaluation results are subsequently converted into reward signals, with event structure constraints incorporated into the reward design to enable iterative optimization of both agents via reinforcement learning.
In three zero-shot scenarios constructed from the RAMS and WikiEvents datasets, our method achieves improvements both in data generation quality and argument extraction performance, while the generated data also effectively enhances the zero-shot performance of other DEAE models.
\end{abstract}

\begin{links}
    \link{Code}{https://github.com/GJZhang2866/GenExtract}
\end{links}

\section{Introduction}
Document-level event argument extraction (DEAE) is one of the essential tasks in information extraction. With the rapid development of large language models (LLMs), DEAE has shown new potential and value in enhancing information retrieval quality, expanding the coverage of knowledge graphs, and enabling trustworthy model editing. Zero-shot document-level event argument extraction (ZS-DEAE) targets the extraction of arguments for unseen event types by utilizing training data associated with seen events. 
The sets of event types used during training and testing are disjoint. However, they are allowed to share overlapping event roles to maintain event structure definition \cite{DBLP:conf/acl/DaganJVHCR18}. Hence, ZS-DEAE aligns with the generalized zero-shot learning paradigm \cite{DBLP:journals/pami/PourpanahALZWLWW23}.

\begin{figure}
    \centering
    \includegraphics[width=\linewidth]{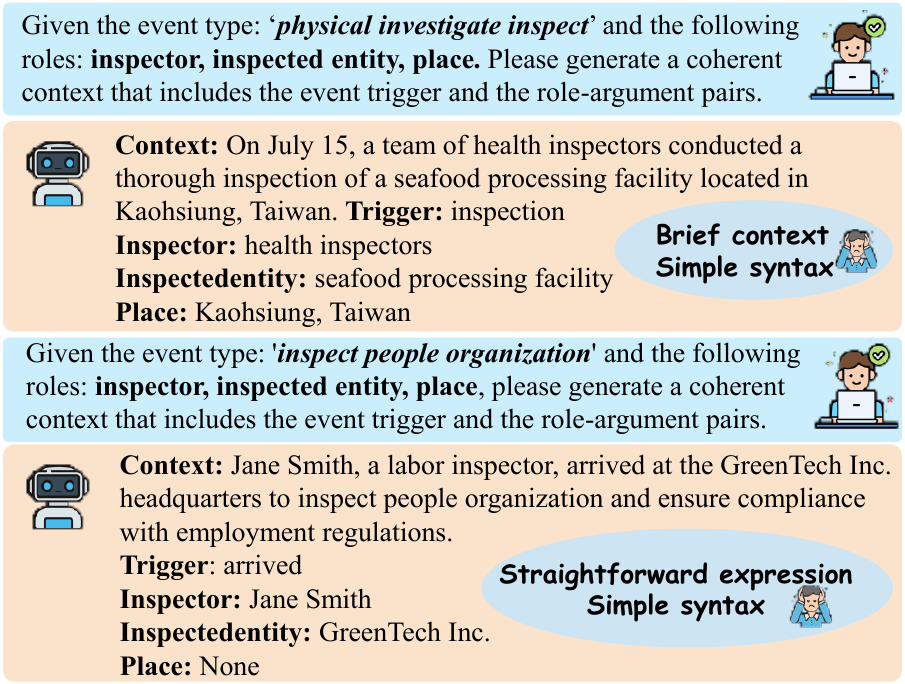}
        \caption{Examples of DEAE data generated by GPT-4o.}
    \label{fig1}
\end{figure}

Most zero-shot event extraction studies focus on sentence-level settings and frame the task as knowledge transfer. Models are typically fine-tuned on seen events using shared semantic spaces or prototypical networks \cite{DBLP:conf/acl/DaganJVHCR18}, enabling limited generalization to unseen types. Another line of work reformulates argument extraction as question answering \cite{eeqa,zstl}, crafting role-specific queries and using pre-trained QA models to identify arguments directly from text.
However, in document-level zero-shot scenarios, these approaches still lag far behind fully supervised models. Empirical evidence shows that DEAE performance improves notably with more annotated data \cite{paie,HMPEAE}, highlighting the continued dependence on high-quality supervision. Although LLMs offer strong reasoning and extensive parametric knowledge \cite{yan2025atomic}, their effectiveness in ZS-DEAE remains weak \cite{sharif2025regen}. This raises a key question: Can LLMs be leveraged to generate structured synthetic data for unseen events to address data scarcity?
Recent efforts explore LLM-generated data for low-resource information extraction \cite{DBLP:conf/emnlp/Zhou0J024,DBLP:conf/acl/XuYZWD24,DBLP:conf/naacl/WangH24}, yet producing high-quality DEAE data is still highly challenging.
Document-level events often involve numerous roles, cross-sentence arguments, and complex semantic relations. This demands that LLMs not only deeply understand event and role semantics, but also generate event instances that are linguistically natural and structurally coherent. Moreover, LLMs often produce factually grounded yet logically incoherent responses\cite{yan2026logicscorefinegrainedlogicevaluation}.  Evaluating the quality of generated data is inherently difficult. 
Without effective quality control, noisy samples can be introduced, which may degrade downstream extraction performance. As illustrated in Fig.\ref{fig1}, the event types ``\textit{inspect people organization}'' and ``\textit{physical investigate inspect}'' both involve the action ``inspect''. The former emphasizes examining people or organizations, while the latter focuses on physical observation via sensory perception or instruments. 
However, LLM-generated examples often fail to clearly distinguish between these event types. 
Additionally, the generated contexts are often linguistically straightforward and arguments that are concentrated, lacking the contextual richness and structural complexity essential for the DEAE.

To address these challenges, we propose a multi-agent collaboration framework that simulates the human cooperative process of “proposal–evaluate–rectify”. The framework comprises two agents: a generation agent that produces document-level contexts with event triggers and structured role–argument pairs based on given event types and roles, and an evaluation agent that extracts arguments and assesses their semantic consistency with the context.
The log-likelihood computed by the evaluation agent serves as a quality indicator and guides subsequent training.
During the generation process, we observe that the generation agent tends to produce instances with multiple empty arguments (roles without corresponding arguments). Although such samples lack valid arguments, the evaluation agent tends to assign them high log-likelihood scores for correctly predicting \texttt{None}, biasing the generation agent toward structurally incomplete events and creating a feedback loop of cumulative bias. 
To mitigate this, we introduce event structural constraints and combine them with log-likelihood into a unified reward signal.  
We employ reinforcement learning to iteratively optimize multi-agent systems, thereby achieving improvements in both data generation quality and argument extraction performance. 
Our contributions are as follows:

\begin{itemize}
    \item We propose a multi-agent collaboration framework for ZS-DEAE to tackle the key challenge of insufficient annotated data for unseen events.
    \item Evaluated on three zero-shot settings derived from RAMS and WikiEvents, our method achieves simultaneous improvements in synthetic data quality and extraction performance.
    \item Our method is generalizable,  enhance the zero-shot capabilities of other models, providing a novel solution for ZS-DEAE.  
\end{itemize}

\section{Related Work}

\textbf{Event argument extraction.} Event argument extraction (EAE) has received growing attention, with datasets such as RAMS \cite{rams} and WikiEvents \cite{bart-gen} driving progress in DEAE. Existing methods include classification-based methods \cite{TSAR,SCPRG}, template-based frameworks \cite{paie,TabEAE,HMPEAE,DEEIA}, and recent LLM-driven models \cite{zhang-etal-2024-ultra,zhou-etal-2024-llms,shuang-etal-2024-thinking}. Despite strong performance, these methods remain heavily dependent on supervised data and show limited generalization to unseen event types.
In zero-shot EAE, early work \cite{DBLP:conf/acl/DaganJVHCR18} maps events into shared neural spaces. QA-based methods were later adapted for unseen argument extraction \cite{eeqa,zstl}. \citet{Dister} further improves generalization by disentangling arguments, roles, and triggers, while other studies enhance zero-shot argument classification using candidate spans \cite{zhang-etal-2021-zero,sainz-etal-2022-textual,lin-etal-2023-global}.

\textbf{Synthetic Data for IE.} 
Synthetic data generation has become a promising strategy for low-resource information extraction. In relation extraction, \citet{zhou-etal-2024-grasping} use LLMs to create seed instances for fine-tuning smaller models and expand pattern diversity with iterative feedback. \citet{DBLP:conf/acl/XuYZWD24} introduce a dual-agent framework to iteratively refine synthetic samples, while \citet{li-etal-2025-generating} integrate in-context learning with direct preference optimization to increase data diversity. For event extraction, \citet{wang-huang-2024-targeted} leverage external corpora to enrich lexical variation in generated samples.

\textbf{Multi-agent for IE.} Multi-agent collaboration has recently emerged as an influential paradigm in IE, spanning cooperative, competitive, and debate-based modes.
\citet{AGENT-NER} propose a cooperative multi-agent system for zero-shot NER that separates entity recognition from type feature extraction and incorporates self-reflection to assess demonstration utility. \citet{zhao-etal-2024-agr,zhao2025learnable} develop agent-guided mechanisms to improve explanation generation. \citet{agent-ee} design a debate-style optimization strategy for few-shot event extraction, and \citet{DBLP:journals/sj/HouJLZW24} present a more scalable multi-agent framework.

\section{Method}
\begin{figure*}
    \centering
    \includegraphics[width=0.9\linewidth]{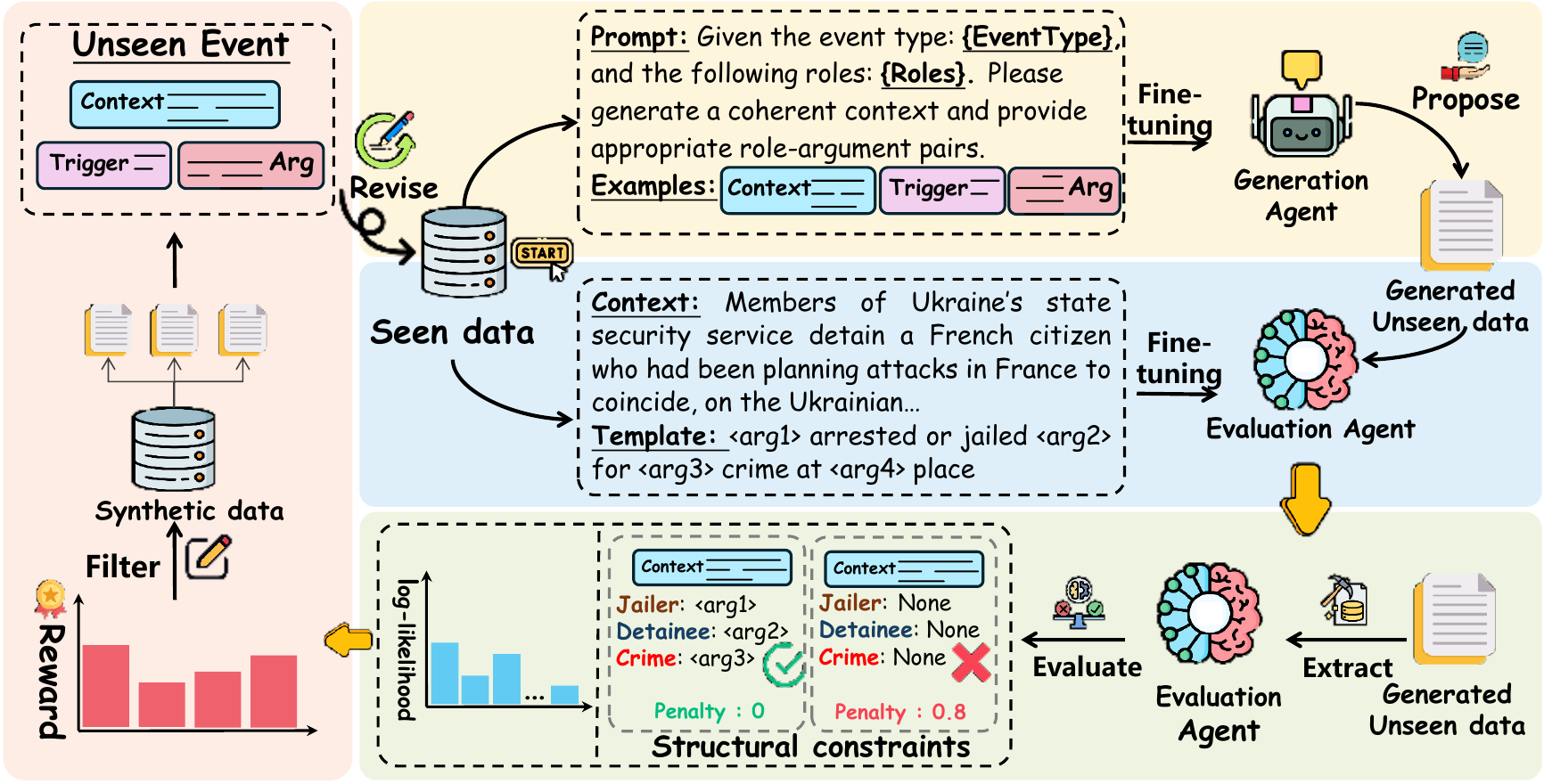}
    \caption{The overall workflow of multi-agent collaboration for ZS-DEAE}
    \label{fig2}
\end{figure*}
\subsection{Task Definition}
\textbf{Document-level Event Argument Extraction (DEAE)} aims to extract textual spans that serve as arguments for specific roles within a given event type, based on the entire input document. Formally, given a document $d$, an event type $e$, a trigger $t$, and a predefined role set $R_e = \{ r_e^i \}_{i=1}^{N_{R_e}}$, the goal of DEAE is to extract a text span corresponding to each role $r_e^i$ from $d$, serving as the argument for that role in the event.

 \textbf{Zero-shot DEAE (ZS-DEAE)} targets the same task without access to labeled data for unseen event types. Specifically, given a set of seen event types $E_s$ and their training dataset $D_s = \{ (d_{e_s}, e_s, t, R_{e_s})_i \}_{i=1}^{|D_s|}$, the goal is to perform argument extraction from documents $D_u$ under unseen event types $E_u$, where $E_s \cap E_u = \emptyset$. Although the event types differ between training and testing, role can be shared between them to ensure the integrity of the event schema.
\subsection{Definition of Agents}
\textbf{Generation Agent.} 
This agent is designed to generate a document-level context $d$, an event trigger $t$, and a set of role-argument pairs $A_e=\{(r_e^i,a_{r_e^i})|r_e^i \in R_e,a_{r_e^i}\in d, i \in [0, N_{R_e}) \}$, conditioned on a given event type $e$ and its corresponding role set $R_e = \{ r_e^i \}_{i=1}^{N_{R_e}}$, where $a_{r_e^i}$ denotes the argument corresponding to role $r_e^i$.

To fully leverage the generation capabilities of LLMs, we formalize the task in a prompt-based input-output format. Specifically, the input prompt is defined as: 
$ E_{\text{in}}(e, R_e) $= `` Given the event type: $e$ and the following roles: $R_e$, please generate a coherent context that includes the event trigger and the role-argument pairs.''  The output is formatted as: $E_{\text{out}}(d, t, A_e)=$`` Context:$d$, Trigger:$t$,Role-Arguments: $A_e$.''
For roles that do not appear in the generated context, their corresponding arguments are marked as \texttt{None}.
The agent is optimized on the training set $D_s$ using an autoregressive objective:

\begin{equation}
\mathcal{L}_g = -\log P(E_{\text{out}} \mid E_{\text{in}}).
\end{equation}

\textbf{Evaluation Agent.}
This agent aims to extract arguments for a given event type $e$ from a document-level context $d$, and to evaluate the structural completeness and semantic consistency of the generated content. In this work, we adopt Bart-Gen\cite{bart-gen} as the evaluation agent, which is based on a conditional generation framework.
The model takes the context $d$ and an unfilled template sentence containing \texttt{<arg>} placeholders as input. The full input is formatted as:
\begin{equation}
c = \texttt{<s>}~\text{template}~\texttt{<s>}~\texttt{</s>}~d~\texttt{</s>},
\end{equation}
and the output is the filled template $x$, where each \texttt{<arg>} placeholder is replaced with an argument extracted from the $d$.
The generation process is modeled autoregressively as generating the target text $x$ token by token, conditioned on both the input $c$ and previously generated tokens:
\begin{equation}
p(x \mid c) = \prod_{i=1}^{|x|} p(x_i \mid x_{<i}, c).
\end{equation}
To reduce hallucination, Bart-Gen restricts its output vocabulary to $V_d$,  containing only tokens present in $d$.
During training, the loss function minimizes the negative log-likelihood over the training set $D_s$:

\begin{equation}
\mathcal{L}_c = -\log P(x \mid c).
\end{equation}

\subsection{Multi-Agent Collaboration}
To address the challenge of limited annotated data in the zero-shot setting, we propose a multi-agent collaborative framework in which two agents work in tandem to simultaneously improve the quality of synthetic data and the performance of event argument extraction. This framework emulates the human collaborative cognitive process of “Propose–Evaluate–Revise”. Specifically, it consists of the following three core stages: (1) Propose: The generation agent attempts to express its understanding of unseen events by generating a document-level context, trigger, and a set of role-argument pairs for a given unseen event type; (2) Evaluate: the evaluation agent performs argument extraction on the generated data and assesses its semantic plausibility and structural completeness; and (3) Revise: the evaluation results are transformed into feedback signals to optimize both agents. Through this iterative and cooperative process, the two agents engage in a dynamic interaction that enables them to improve together over time.  The overall
workflow is described in Fig. \ref{fig2}.

\textbf{Propose.} Given an unseen event type $e_{u_i}$ and its associated roles $R_{e_{u_i}}$, the generation agent expresses its understanding by generating document-level contexts, an event trigger, and role-argument pairs. Specifically, the event and roles are first mapped into a natural language prompt $E_{\text{in}}(e_{u_i}, R_{e_{u_i}})$, and the generation agent produces $K$ candidate sequences based on this prompt. During generation, some contexts may omit the specified trigger or certain arguments. Therefore, samples missing the trigger in the generated context are discarded, while missing arguments are assigned the value \texttt{None}.
Each valid  generated sequence is decoded into a synthetic data instance represented as the tuple $(d, t, A_{e_{u_i}})$. Collecting all $K$ samples yields   synthetic dataset for $e_{u_i}$ denoted as $D_{\text{syn}}^{e_{u_i}} = \{ (d_j, t_j, A_{e_{u_i}}^{j}) \}_{j=1}^{K}$.
The complete synthetic dataset for all unseen event types is formed by the union $D_{\text{syn}} = D_{\text{syn}}^{e_{u_1}} \cup \dots \cup D_{\text{syn}}^{e_{u_m}}$.

\textbf{Evaluate.} After the generation agent expresses its understanding of the event, we use the evaluation agent to assess whether this understanding is accurate. The generation probability $P(x \mid c)$ of the evaluation agent can be interpreted as the likelihood of generating a specific argument-filled sentence $x$ given the input $c$, or as the semantic match between c and x, which can then be used to measure the quality of the synthetic data $(d, t,A_{e_{u_i}}^{j}) \in D_{\text{syn}}^{e_{u_i}}$. Specifically, for each synthetic sample, we compute the log-likelihood of its corresponding filled template:
\begin{equation}
\ell_i = \log P(x_i \mid c_i).
\end{equation}
High-quality samples are expected to yield higher log-likelihoods, while lower-quality samples should have lower scores, thereby enabling the differentiation of synthetic data quality.
To further standardize this scoring, we normalized the log-likelihood:
\begin{equation}
\alpha_i = \frac{\ell_i - \mu}{\delta},
\end{equation}
where $\mu$ and $\delta$ denote the mean and standard deviation of log-likelihood values over the entire synthetic dataset $D_{\text{syn}}$.

During evaluation, we observed that many synthetic samples contain multiple empty arguments, meaning arguments that are generated as \texttt{None}. Such samples often receive high likelihood scores from the evaluation agent as it correctly predicts these \texttt{None}. This tendency may further induce the generation agent to prefer producing structurally incomplete events, thereby introducing and amplifying bias in the feedback loop. 
To mitigate this issue, we introduce the structural completeness constraint by penalizing the proportion of empty arguments in each synthetic sample. Specifically, for each sample, we compute the proportion of empty arguments $\rho_i$. We aim to keep this proportion close to that observed in the training data, whose expected value and standard deviation are denoted by $\tau$ and $\varepsilon$, respectively. The penalty term is defined as:

\begin{equation}
\text{p}_i =
\begin{cases}
0 & \text{if } \tau - \varepsilon \leq \rho_i \leq \tau + \varepsilon \\
|\rho_i - \tau| & \text{otherwise}
\end{cases}
\end{equation}
Finally, we integrate this penalty into the normalized log-likelihood score to obtain the final quality score for each synthetic sample:
\begin{equation}
\alpha_i = \frac{\ell_i - \mu}{\delta} - \text{p}_i.
\end{equation}

\textbf{Revise.} Reinforcement learning (RL) has been shown to be effective in learning preferences among data instances. To this end, we adopt an RL-based approach to refine both agents. By assigning higher rewards to higher-quality synthetic samples, the two agents are encouraged to better understand unseen events through data augmentation, thereby achieving joint improvement in both synthetic data quality and argument extraction performance.
Specifically, we treat the normalized score $\alpha_i$ as the reward and optimize the generation agent and evaluation agent via policy gradient methods based on the expected reward over the synthetic dataset $D_{\text{syn}}$. The parameter updates for the two agents are defined as:

\begin{equation}
\mathcal{G}_{i+1} = \mathcal{G}_{i} + \gamma_1 \nabla_\mathcal{G} \mathbb{E}[\alpha],
\end{equation}
\begin{equation}
\mathcal{E}_{i+1} = \mathcal{E}_{i} + \gamma_2 \nabla_\mathcal{E} \mathbb{E}[\alpha],
\end{equation}

where $\gamma_1$ and $\gamma_2$ are the learning rates for the generation agent and the evaluation agent, respectively. The gradients are computed as:
\begin{equation}
\nabla_G \mathbb{E}[\alpha] = \mathbb{E}[\alpha_i \nabla_\mathcal{G} \log P(E_{\text{out}} \mid E_{\text{in}})],
\end{equation}
\begin{equation}
\nabla_C \mathbb{E}[\alpha] = \mathbb{E}[\alpha_i \nabla_\mathcal{E} \log P(x \mid c)].
\end{equation}

Through continuous cycles of propose-evaluate-revise, the two agents progressively enhance their understanding of unseen events, effectively distinguish subtle semantic differences between similar event types, and iteratively improve their respective capabilities.

\section{Experiments}
\begin{table*}[ht]
	\centering
        \small
	\begin{tabular}{lccccccccc}
		\toprule
		\textbf{Model} & \multicolumn{3}{c}{\texttt{RAMS2RAMS}} & \multicolumn{3}{c}{\texttt{RAMS2Wiki}} & \multicolumn{3}{c}{\texttt{Wiki2Wiki}} \\
		\cmidrule(lr){2-4}\cmidrule(lr){5-7}\cmidrule(lr){8-10}
		& {Seen R.} & {Unseen R.} & {Overall} &  {Seen R.} & {Unseen R.} & {Overall} & {Seen R.} & {Unseen R.} & {Overall} \\
		\bottomrule
		\multicolumn{10}{c}{{DEAE Models}} \\
		\toprule
		{PAIE} & 32.52 & 28.87 & 30.80  & 19.57& 31.72  & 20.15 &23.58 &23.57 &24.42 \\
		{TabEAE} & 37.16 & 35.26 & 36.22  & 16.94 & \textbf{35.05} & 26.74 & 37.19 &28.84 &30.97 \\
		{DEEIA} &36.57	&39.49	&37.95	&1.50&7.17&5.12	&34.11	&19.48	&22.51 \\
		{HMPEAE} &35.18	&37.74	&36.44		&16.89 &32.74	&25.61	&38.43	&27.48	&30.20 \\
            {TSAR} &38.10	 &21.56&	30.90	&15.77		&13.37 &11.71	&14.40&	13.86	&13.95 \\
		{SCPRG} &38.93	&26.97	&33.58	&10.80	&10.00 &9.40		&45.80	&11.89	&21.90\\
		\bottomrule
		\multicolumn{10}{c}{{Zero-shot Models}} \\
		\toprule
            EEQA	&26.75	&26.99	&26.86	&14.74	&21.39&	18.24&	23.05	&14.00	&21.50 \\
            ZSTL	&-	&-	&12.24	&-	&-	&9.40	&-	&-	&9.97 \\
		{Bart-Gen} &39.89	&37.09	&38.53	&24.66	&33.45	&28.52	&48.11	&32.68	&40.82\\
		{Distar} 	&18.26	&19.68	&18.98	&	13.11 &3.20	&10.60	&2.72	&20.18	&17.51 \\
		
            \bottomrule
		\multicolumn{10}{c}{{Large Language Models}} \\
		\toprule
            \includegraphics[width=0.25cm, height=0.25cm]{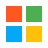}{Phi-4} 	&-	&-	&10.10	&-	&- &4.12	&-	&- &7.30 \\
            \hspace{2em} +\texttt{CoT}			&-	&-	&12.89	&-	&-	&2.78 &-	&-	&5.78				\\
            \includegraphics[width=0.25cm, height=0.25cm]{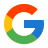} {Gemma-1.1} 	&-	&-	&13.39	&-	&-	&5.91	&-	&-	&6.31\\
		\hspace{2em} +\texttt{CoT}			&-	&-	&5.94	&-	&-	&3.19	&-	&-	&4.78 		\\
            \includegraphics[width=0.25cm, height=0.25cm]{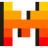} {Mixtral}  &-	&-	&19.35	&-	&-	&4.70	&-	&-	&6.34\\
    		\hspace{2em} +\texttt{CoT}			&-	&-	&15.96	&-	&-	&4.81	&-	&-	&6.23 		\\
		
		\includegraphics[width=0.25cm, height=0.25cm]{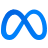} {LLaMA-3.1}  	 &-	&-	&20.56	 &-	&-	&8.37	 &-	&-	&10.55	 	\\
		\hspace{2em} +\texttt{CoT}			 &-	&-	&17.05	 &-	&-	&6.78	 &-	&-	&8.87		\\
		
		\includegraphics[width=0.25cm, height=0.25cm]{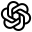} { GPT-4o}  &-	&-	&19.64	&-	&-	&10.17	&-	&-	&11.36		\\
		\hspace{2em} +\texttt{CoT}			&-	&-	&18.36	&-	&-	&7.98 &-	&-	&9.65				\\

		\includegraphics[width=0.30cm, height=0.25cm]{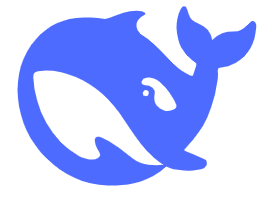} { DS-V3}  &-	&-	&22.00	 &-	&-	&10.89	 &-	&-	&11.60		\\
		\hspace{2em} +\texttt{CoT}			 &-	&-	&15.24	 &-	&-	&7.63 &-	&-	&8.62			\\

		\includegraphics[width=0.30cm, height=0.25cm]{figs/icons/deepseek.png} { DS-R1} &-	&-	&24.41	 &-	&-	&12.27  &-	&-	&12.84	\\
		\hspace{2em} +\texttt{CoT}		 &-	&-	&23.47		 &-	&-	&9.24 &-	&-	&10.40			\\
		\hline
		Ours (LLaMA)	&\textbf{46.46}	&\underline{45.06}	&\textbf{45.77}	&\textbf{30.81}	&\underline{34.43}	&\textbf{32.38}	&\underline{47.83}	&\underline{46.19}	&\underline{46.96}\\
    Ours (Qwen)	&\underline{44.06}	&\textbf{45.11}	&\underline{44.59}	&\underline{ 31.74}	&{30.47}	&\underline{31.18}	&\textbf{47.39}	&\textbf{47.82}	&\textbf{47.62} \\
        \bottomrule
	\end{tabular}
    
	\caption{Main Results on three zero-shot settings. Seen R. refers to roles shared between seen and unseen event schemas, while Unseen R. denotes roles unique to unseen schemas. Overall indicates the F1 score calculated over all roles. We use \textbf{bold} text to indicate the best performance and \underline{underline} to indicate the second best. For ZSTL and LLM-based methods, since they are not trained on $D_s$ but instead make predictions directly via queries or prompts, they do not distinguish between seen and unseen roles—thus, the corresponding entries are left blank. }
	\label{main}
\end{table*}
\subsection{ Experimental Settings}

\textbf{Datasets.} We construct three zero-shot settings based on the two most widely used DEAE datasets, RAMS and WikiEvents. These settings are \texttt{RAMS2RAMS}, \texttt{RAMS2Wiki}, and \texttt{Wiki2Wiki}.
In the \texttt{RAMS2RAMS},  the RAMS dataset is split into seen and unseen event types.  For \texttt{RAMS2Wiki}, we remove overlapping event types in WikiEvents that appear in RAMS, ensuring the test set contains only disjoint event types. The \texttt{Wiki2Wiki} setting follows the same protocol as \texttt{RAMS2RAMS}, where WikiEvents is split into seen and unseen event types.

\textbf{Experimental Configuration and Metrics} 
We select LLaMA3.1-8B \cite{llama3} and Qwen2.5-7B \cite{Qwen2.5} as generation agent, fine-tuned via LoRA \cite{lora} for parameter-efficient adaptation.  LoRA modules are inserted into the attention layers (query and value projections), with rank 8, scaling factor 32, and dropout rate 0.05.
For the evaluation agent, we use Bart-large \cite{bart} as the pre-trained model for Bart-Gen \cite{bart-gen}. We run the five-round agent interaction optimization with three different random seeds, and report the average result of the best-performing round from each run, to reduce the impact of randomness.
For evaluation metrics, we adopt Span-F1 as the primary metric, where a prediction is correct only if the span exactly matches the gold span.

\textbf{Baselines.} We compare our method with three categories of baselines. The first includes \textbf{DEAE models}, such as PAIE \cite{paie}, TabEAE \cite{TabEAE}, DEEIA \cite{DEEIA}, HMPEAE \cite{HMPEAE}, TSAR \cite{TSAR}, and SCPRG\cite{SCPRG}. Since TSAR and SCPRG are not designed for zero-shot settings, we adapt them into prototype-based classifiers for transfer to unseen event types.
The second includes \textbf{sentence-level zero-shot models}, including EEQA \cite{eeqa}, ZSTL\cite{zstl}, Bart-gen \cite{bart-gen}, and Distar\cite{Dister}. We adapt EEQA and ZSTL to our task format by redesigning their prompt templates.
The third includes \textbf{LLMs}, such as Phi-4 (14B) \cite{phi4}, Gemma-1.1 (7B) \cite{Gemma}, Mixtral (8x22B) \cite{Mixtral}, LLaMA3.1 (70B) \cite{llama3}, GPT-4o\cite{DBLP:journals/corr/abs-2303-08774}, DeepSeek V3 (DS-V3) \cite{DS-V3} and DeepSeek R1 (DS-R1) \cite{DS-R1}. We evaluate them using both zero-shot and Chain-of-Thought (CoT) prompting following \citet{sharif2025regen}.
\subsection{Main results}

The main results are shown in Table \ref{main}.
\textbf{ Comparison with DEAE Models.}
Our method, based on LLaMA and Qwen, consistently outperforms existing DEAE models across all three zero-shot settings. Ours (LLaMA) exceeds the strongest baseline DEEIA by 6.57, 5.57, and 7.82 F1 on seen roles, unseen roles, and overall in \texttt{RAMS2RAMS}, and achieves the highest overall F1 scores of 32.38 and 46.96 in \texttt{RAMS2Wiki} and \texttt{Wiki2Wiki}. These results reveal the limitations of prior DEAE models in zero-shot scenarios, as their dependence on manually annotated data reduces generalization. Our method also achieves more balanced performance between seen and unseen roles.

\textbf{Compared with zero-shot models}. Our method shows substantial gains over Bart-Gen, with overall F1 improvements of 7.53, 3.86, and 6.14 across datasets, benefiting from collaborative interaction between generation and evaluation agents. In contrast, EEQA, ZSTL, and Dister perform poorly, indicating limited capacity in document-level settings.
\textbf{Comparison with LLMs.} We evaluate several mainstream LLMs using both Zero-shot and CoT prompts. The overall results show that LLMs perform significantly worse in Span-F1, with generally low performance levels. This aligns with observations by \citet{sharif2025regen}. A major factor is the strict boundary matching required by Span-F1. Although LLMs demonstrate strong language understanding, they often struggle to accurately identify precise argument spans. More flexible evaluation metrics may better reflect their potential.

\subsection{Ablation Study}
\begin{table}[t]
	\centering
	\small
	\begin{tabular}{lccccccc}
		\toprule
		& \multicolumn{1}{c}{Model}&   \multicolumn{1}{c}{\textbf{R2R}} & \multicolumn{1}{c}{\textbf{R2W}} & \multicolumn{1}{c}{{\textbf{W2W}}} \\
		\hline
		&Ours((LLaMA)		&\textbf{45.28}	&\textbf{32.38}	&\textbf{46.96}	\\
		&-reward	 		&42.46	&24.93	&38.39	\\
		&-constraint 		&44.72	&29.03	&40.98 	\\
        \hline
		&Ours(QWEN)			&\textbf{44.59}	&\textbf{31.18}	&\textbf{47.62}	\\
		&-reward	 		&40.52	&27.19	&43.30	\\
		&-constraint 		&40.87	&27.87	&38.70	\\
		\bottomrule
	\end{tabular}
	\caption{Ablation study on the effects of reward and constraint across three datasets. \textbf{R2R}: \texttt{RAMS2RAMS}. \textbf{R2W}: \texttt{RAMS2Wiki}, \textbf{W2W}: \texttt{Wiki2Wiki}}
	\label{Ablation}
\end{table}

\begin{table*}[htbp]
\small
\centering
\begin{tabular}{l|l|ccc}
    \hline
    \textbf{Model} & \textbf{Data} & \texttt{RAMS2RAMS} & \texttt{RAMS2Wiki} & \texttt{Wiki2Wiki} \\
    \hline
    {TabEAE} 
        & Seen data             & 36.22 & 26.74 & 30.97 \\
        & + Synthetic data ((LLaMA 3.1) & 37.37 & 26.17 & 12.48 \\
        & + Synthetic data (Ours)    & \textbf{44.43 }& \textbf{28.42}& \textbf{33.36} \\
    \hline
    {Bart-Gen} 
        & Seen data             & 38.53 & 28.52 & 40.82 \\
        & + Synthetic data (LLaMA 3.1) & 41.04 & 28.95 & 44.17 \\
        & + Synthetic data (Ours)    & \textbf{46.06} & \textbf{32.38} & \textbf{46.96} \\
    \hline
    \end{tabular}
    \caption{Performance comparison on three transfer settings using unseen and synthetic data.}
    \label{tab:synthetic-data-quality}
\end{table*}

\begin{figure}
    \centering
    \includegraphics[width=0.88\linewidth]{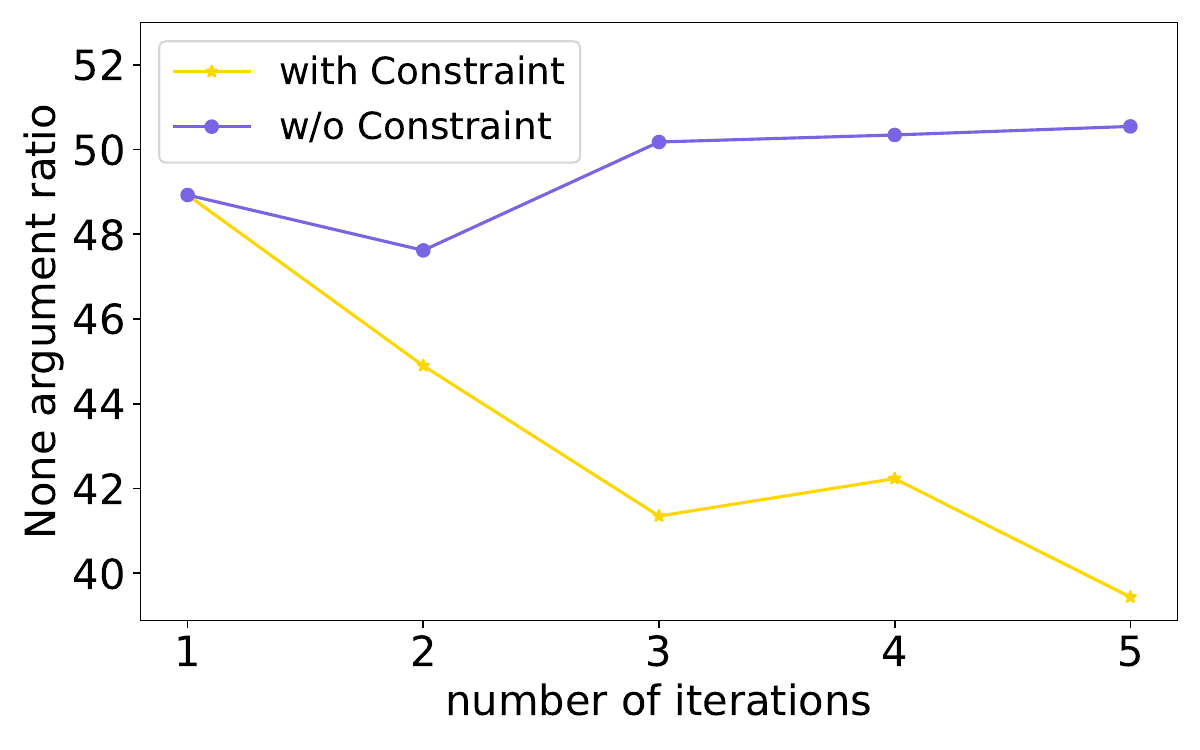}
    \caption{Proportion of empty arguments in synthetic data. }
    \label{fig3}
\end{figure}

Our framework consists of the RL feedback mechanism and the event structure constraint. To evaluate their contributions, we conduct ablation studies by removing each component, as shown in Table \ref{Ablation}. Removing either the RL reward or the structure constraint reduces performance, which shows that both components are effective.
In most settings, removing the structure constraint produces a smaller performance drop, which suggests that multi-agent collaboration is the main factor. However, Fig.\ref{fig3} shows that the absence of this constraint increases the proportion of empty arguments and harms both data quality and extraction accuracy. In contrast, applying the constraint reduces empty arguments and improves the completeness of arguments as well as the quality of synthetic data.
\section{Analysis}
\subsection{Impact of Interaction Rounds}
\begin{figure}[t]
	\centering
 
	\begin{subfigure}{0.48\linewidth}
		\includegraphics[width=1\linewidth]{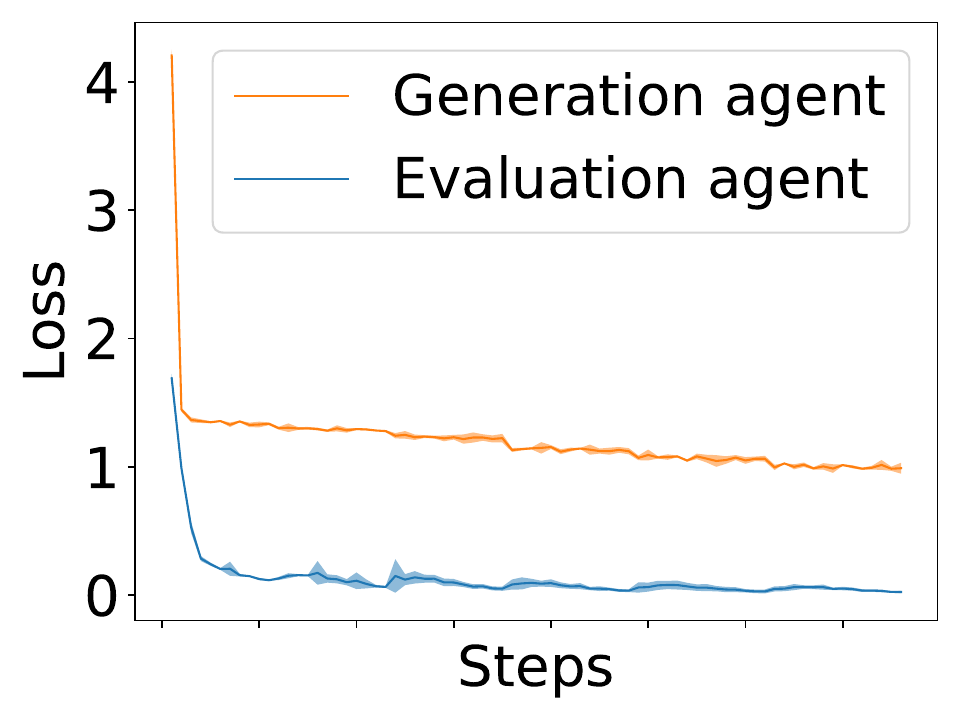}
		\caption{}
		\label{h_a}
	\end{subfigure}
	\begin{subfigure}{0.47\linewidth}
		\includegraphics[width=\linewidth]{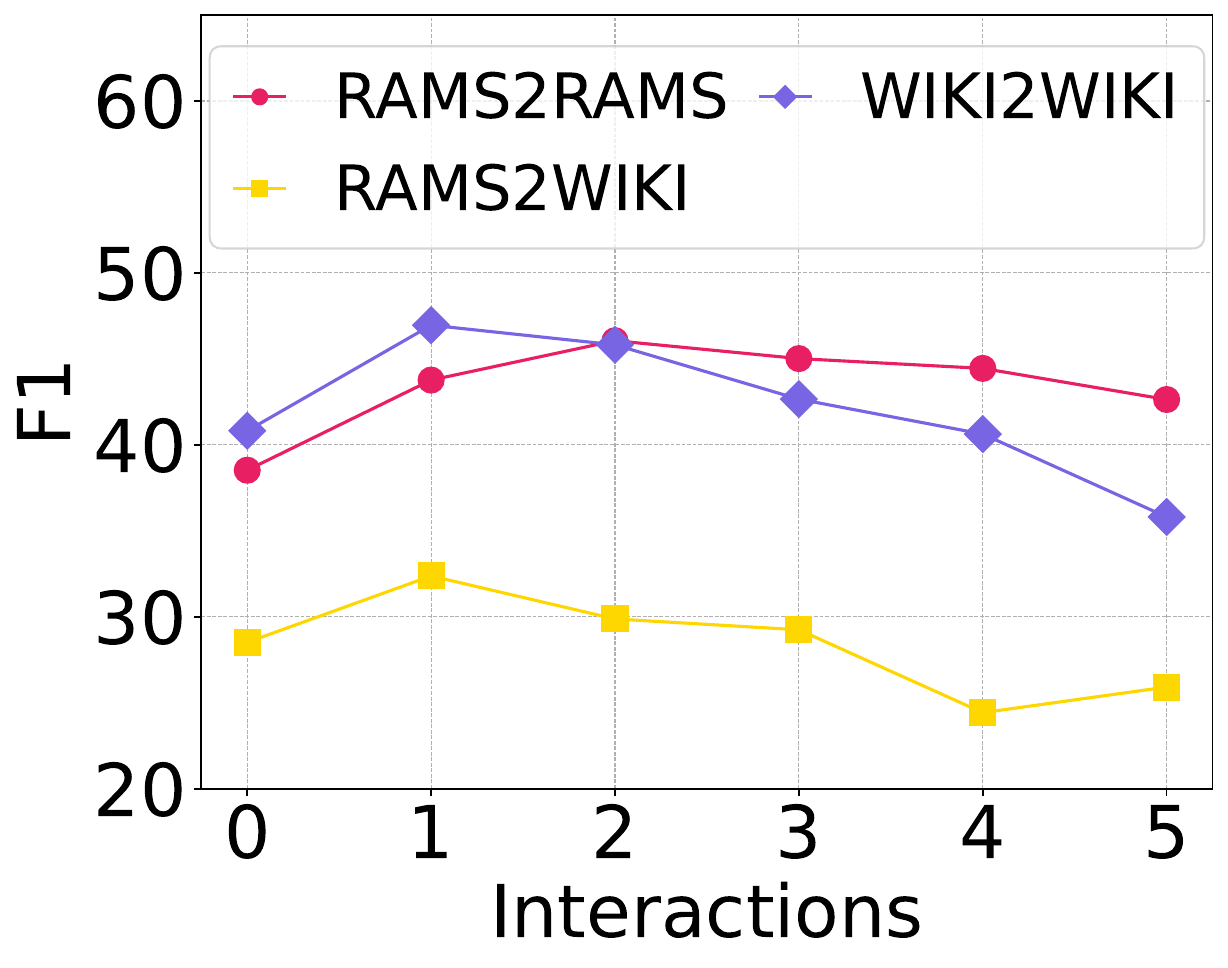}
		\caption{}
		\label{h_b}
	\end{subfigure}

	\caption{(a): Loss of the two agent during reinforcement learning on \texttt{RAMS2RAMS}. (b): F1 trend across three datasets.}
	\label{Hyperparameter}
\end{figure}

Fig. \ref{h_a} shows the loss curves of the generation and evaluation agents, assessing RL process stability. Both losses decline steadily during training, indicating that the optimization method supports cooperative convergence and ensures stable training.
Fig. \ref{h_b} presents F1 trends across three dataset settings. The model reaches peak F1 scores within one to two interaction rounds, but performance gradually declines with more interactions.
As analyzed later, this may result from reduced diversity in generated samples, which undermines the model’s generalization.

\subsection{Analysis of Generation Agent}
To evaluate the quality of synthetic data generated by the Generation Agent, we augment the training sets of baseline models and compare them with data generated directly by LLaMA under the same settings. Performance improvements from each source serve as an indirect measure of data quality.We conduct experiments on two representative DEAE models: TabEAE, which shows balanced performance (Table \ref{main}), and Bart-Gen, a stronger zero-shot model. As shown in Table \ref{tab:synthetic-data-quality}, our method consistently achieves stable and significant gains across different settings, outperforming LLaMA-generated data. These results indicate that our synthetic data is higher in quality and generalizability, effectively enhancing ZS-DEAE.

\subsection{Analysis of Evaluation Agent}
\begin{figure}
    \centering
    \includegraphics[width=0.92\linewidth]{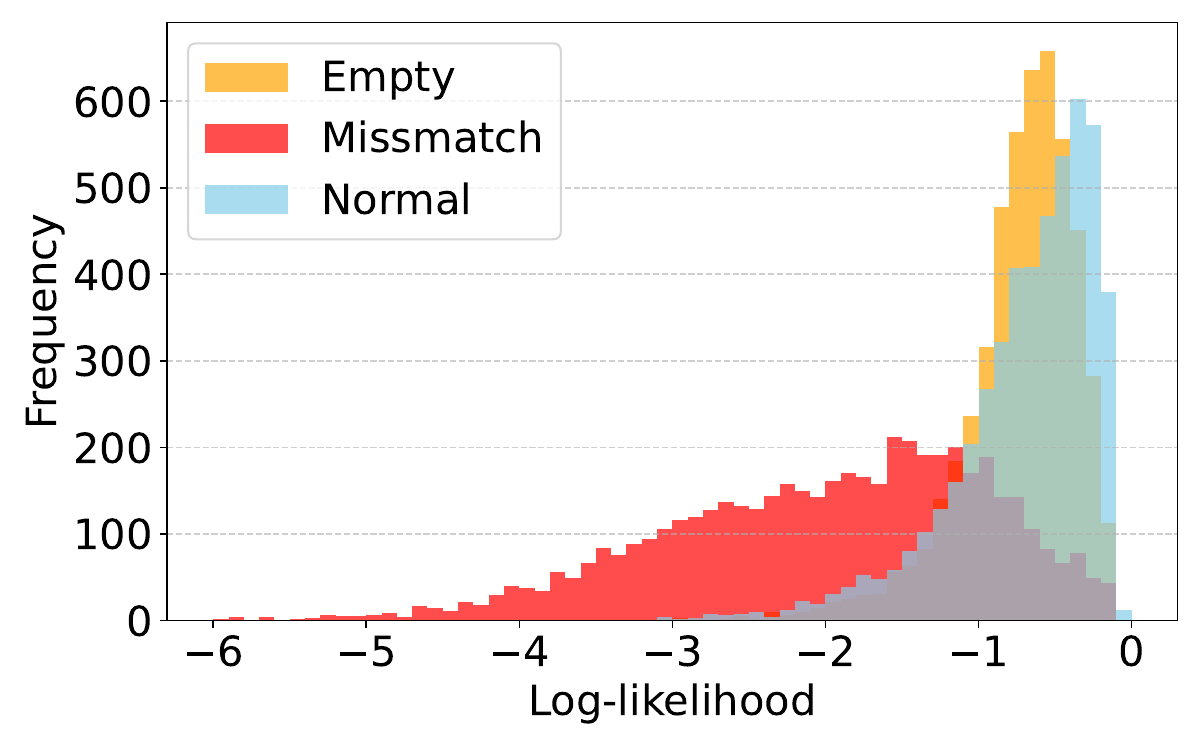}
    \caption{Log-likelihood scores assigned by the evaluation agent to different types of synthetic data. Empty: randomly removing arguments. Mismatch: randomly swapping argument-role mappings. Normal: correctly generated data without perturbation.}
    \label{fig4}
\end{figure}

To assess the evaluation agent’s ability to distinguish different data quality, we construct low-quality variants of the test set by randomly replacing or removing arguments.Each instance is scored using log-likelihood, with results shown in Fig.\ref{fig4}. consistently assigns higher log-likelihoods to high-quality data and lower scores to degraded examples. This indicates that the agent possesses a certain degree of sensitivity to data quality and can effectively differentiate between well-formed and degraded examples.
\subsection{Diversity Analysis of Synthetic Data}
\label{diver}
\begin{figure}[t]
	\centering
	\begin{subfigure}{0.49\linewidth}
		\includegraphics[width=\linewidth]{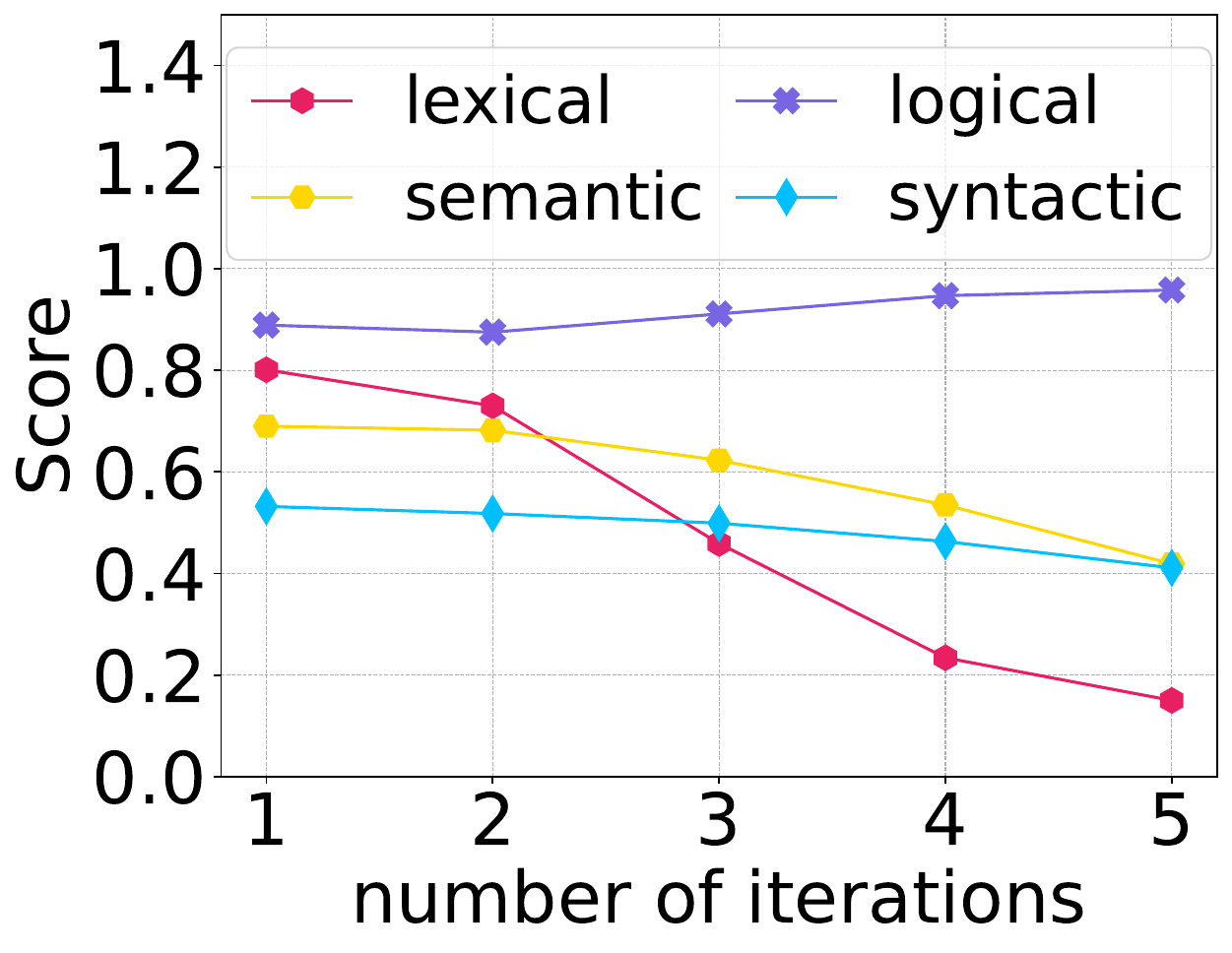}
		\caption{Per-input diversity}
		\label{d_1}
	\end{subfigure}
	\begin{subfigure}{0.49\linewidth}
		\includegraphics[width=\linewidth]{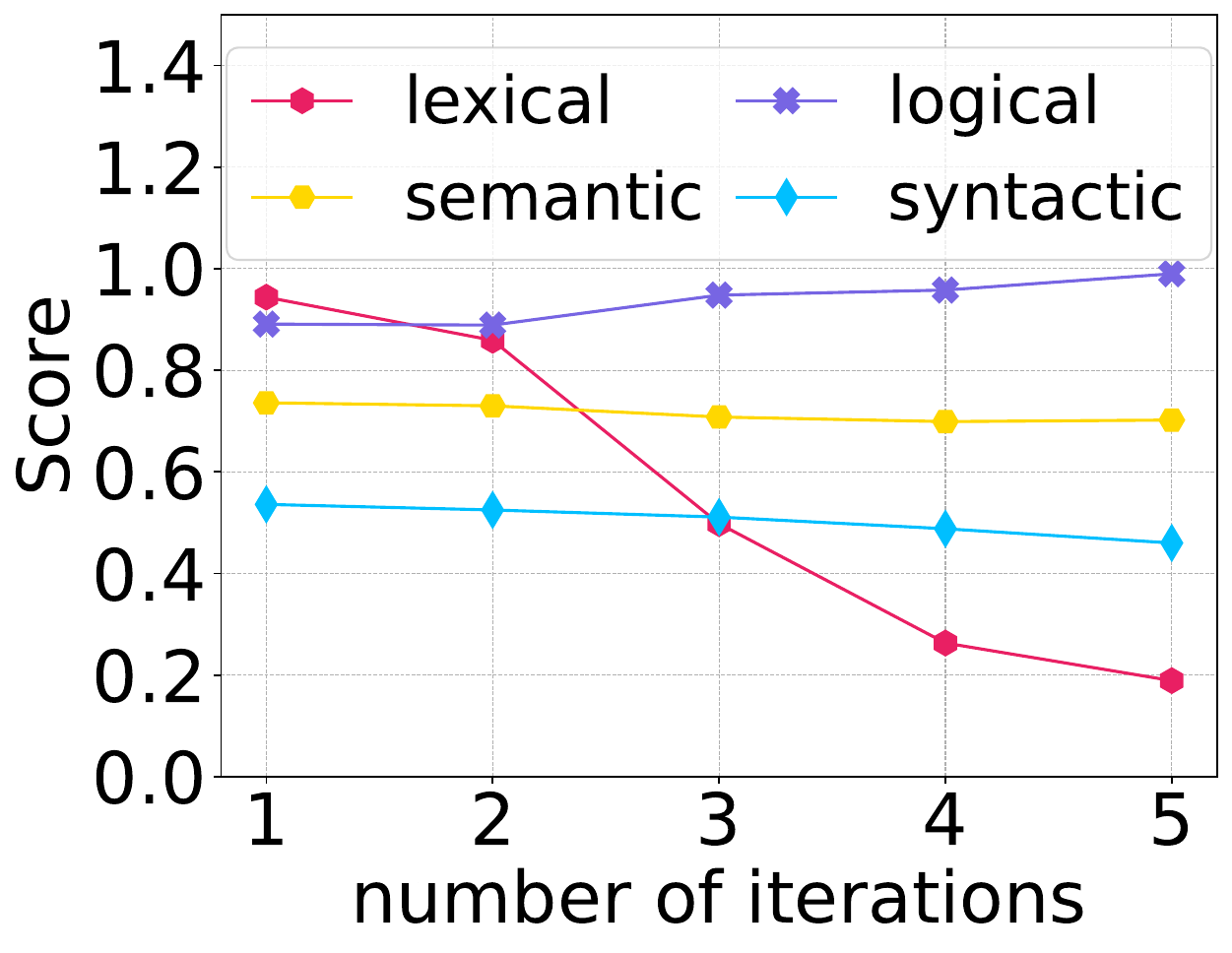}
		\caption{Across-input diversity}
		\label{d_1}
	\end{subfigure}
	\caption{Diversity of Synthetic Data.}
	\label{Diversity}
\end{figure}

We evaluate synthetic data diversity following \citet{DBLP:conf/iclr/KirkMNLHGR24}, which separates per-input and across-input diversity, and adopt the four-dimensional analysis of \citep{DBLP:journals/corr/abs-2412-10271}: lexical, semantic, logical, and syntactic. Results appear in Fig.\ref{Diversity}.
As interaction rounds increase, diversity decreases in the lexical, semantic and syntactic dimensions, indicating that the model shifts toward stable generation strategies that favor patterns with higher log-likelihood. This convergence may improve consistency but reduces generalization and contributes to performance decline in later rounds. Logical-level diversity remains relatively stable, possibly because lexical and syntactic convergence introduces inconsistencies that sustain variation in logical content. Future work should enhance diversity, especially in the lexical, semantic and syntactic.
\subsection{Case Study}
\begin{figure}
    \centering
    \includegraphics[width=\linewidth]{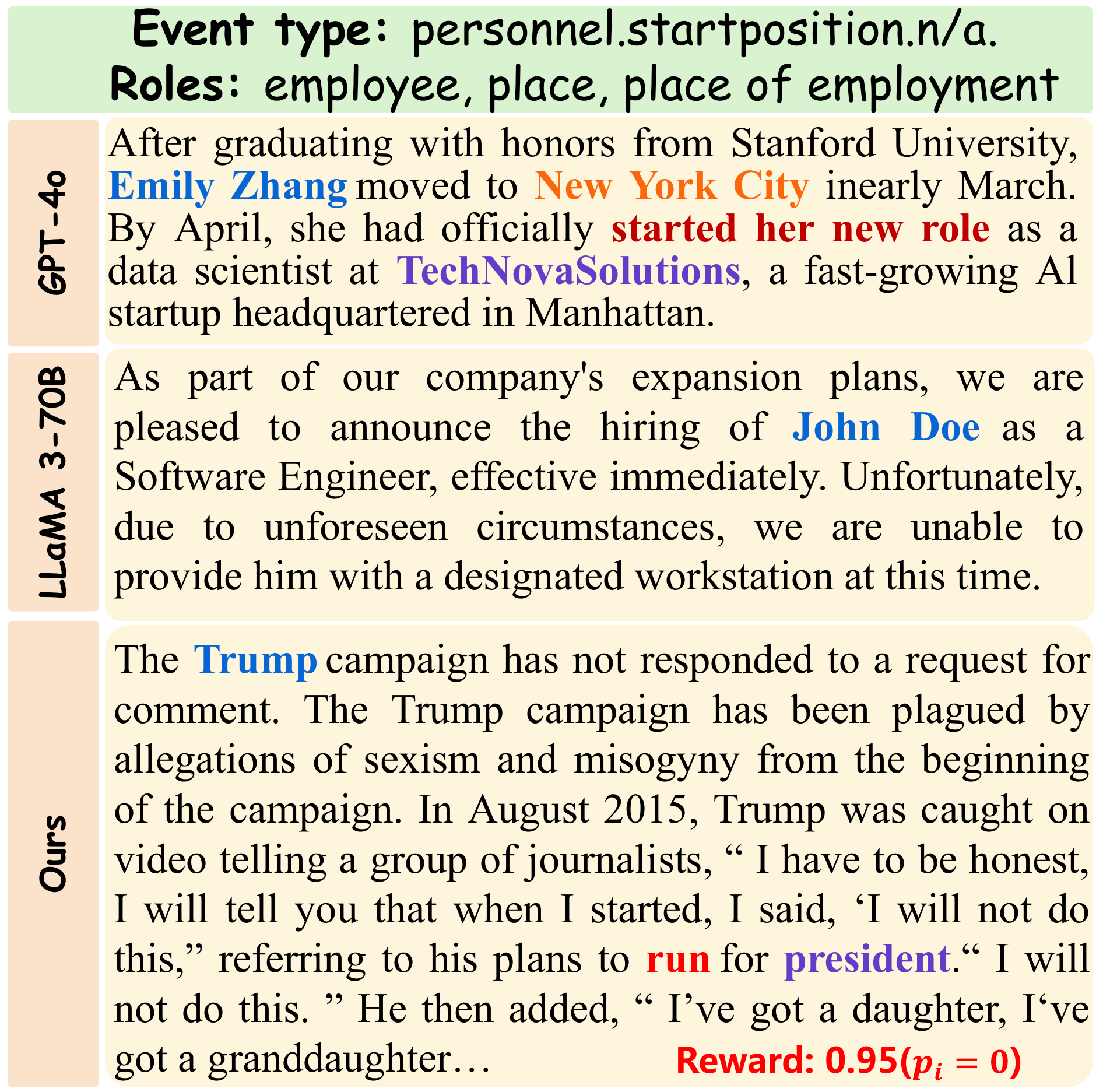}
    \caption{Case studies of synthetic data.}
    \label{case}
\end{figure}
We present a case study in Fig.\ref{case}, analyzing synthetic data generated by GPT-4o, LLaMA3.1 (70B), and our method using LLaMA3.1 (8B). The outputs from GPT-4o and LLaMA3.1 (70B) tend to be short, syntactically simple, and contains densely packed arguments, making them less suitable for DEAE. Notably, LLaMA3.1 (70B) often omits the trigger, as well as key arguments of \textit{place} and \textit{placeofemployment}. In contrast, our method generates a longer document with more widely dispersed arguments. Moreover, in the fourth case, a mismatch between the context and event type results in a low reward, highlighting the effectiveness of evaluation agent.

\section{Conclusion}
This paper presents a multi-agent collaborative framework for ZS-DEAE that mitigates the shortage of annotated data. The framework models a “Propose–Evaluate–Revise” cycle to iteratively improve synthetic data and enhance extraction accuracy. Experiments show clear improvements in data quality and argument extraction, surpassing several mainstream LLMs. Additional analyses indicate that the generated data can also strengthen other models in zero-shot settings, providing a promising solution for ZS-DEAE. Future work will extend this framework to broader information extraction tasks and other low-resource scenarios.
\section{Acknowledgments}
We thank all the anonymous reviewers for their constructive comments and suggestions. This work is supported by the National Natural Science Foundation of China (62176145, 62476161), and the Interdisciplinary Research Fund of Shanxi University.
\bibliography{aaai2026}

\end{document}